\pdfoutput=1

\documentclass[11pt]{article}

\usepackage[final]{latex/acl}

\usepackage{times}
\usepackage{latexsym}

\usepackage[T1]{fontenc}

\usepackage[utf8]{inputenc}

\usepackage{microtype}

\usepackage{inconsolata}

\usepackage{graphicx}

\usepackage{booktabs}
\usepackage{multirow}
\usepackage{graphicx}

\usepackage{pifont}
\usepackage{amsmath}
\usepackage{amssymb}
\usepackage{ifthen}
\usepackage{dsfont}

\usepackage{listings}
\usepackage{xcolor}
\usepackage{ltablex}
\usepackage{longtable}
\usepackage{multicol}

\usepackage{tikz}
\usetikzlibrary{bayesnet}
\usepackage{subfloat}
\usepackage{subfig}
\usepackage[linguistics]{forest}

\usepackage[ruled,vlined]{algorithm2e}

\usepackage{caption}
\usepackage{subcaption}

\usepackage{cancel}

\usepackage{xspace}
\usepackage[inline,shortlabels]{enumitem}

\usepackage[noabbrev,capitalize]{cleveref}

\newcommand{\prob}[2][]{p\ifthenelse{\not\equal{}{#1}}{_{#1}}{}(#2)} 
\newcommand{\expect}[2][]{\text{\bf E}\ifthenelse{\not\equal{}{#1}}{_{#1}}{}\!\left[#2\right]}
\newcommand{\var}[2][]{\text{\bf Var}\ifthenelse{\not\equal{}{#1}}{_{#1}}{}\!\left[#2\right]}

\title{Context-Efficient Retrieval with Factual Decomposition}

\author{Yanhong Li \\
  University of Chicago / TTIC \\\texttt{yanhongli@uchicago.edu} \\\And
  David Yunis \\
  Toyota Technological Institute at Chicago \\
  \texttt{dyunis@ttic.edu} 
  \\\AND
  David McAllester \\
  Toyota Technological Institute at Chicago\\\texttt{mcallester@ttic.edu} \\\And
  Jiawei Zhou \\
  Stony Brook University\\\texttt{jiawei.zhou.1@stonybrook.edu} \\}

\begin{document}
\maketitle
\begin{abstract}

There has recently been considerable interest in incorporating information retrieval into  large language models (LLMs).  Retrieval from a dynamically expanding external corpus of text allows a model to incorporate current events and can be viewed as a form of episodic memory.
Here we demonstrate that pre-processing the external corpus into semi-structured ``atomic facts'' makes retrieval more efficient. More specifically, we demonstrate that our particular form of atomic facts improves performance on various question answering tasks when the amount of retrieved text is limited. Limiting the amount of retrieval reduces the size of the context and improves inference efficiency.\footnote{Code available at \url{https://github.com/yanhong-lbh/DecompFacts}.}

\end{abstract}

\section{Introduction}
\label{sec-intro}

\begin{figure}[h!]
    \centering
    \includegraphics[width=0.9\linewidth]{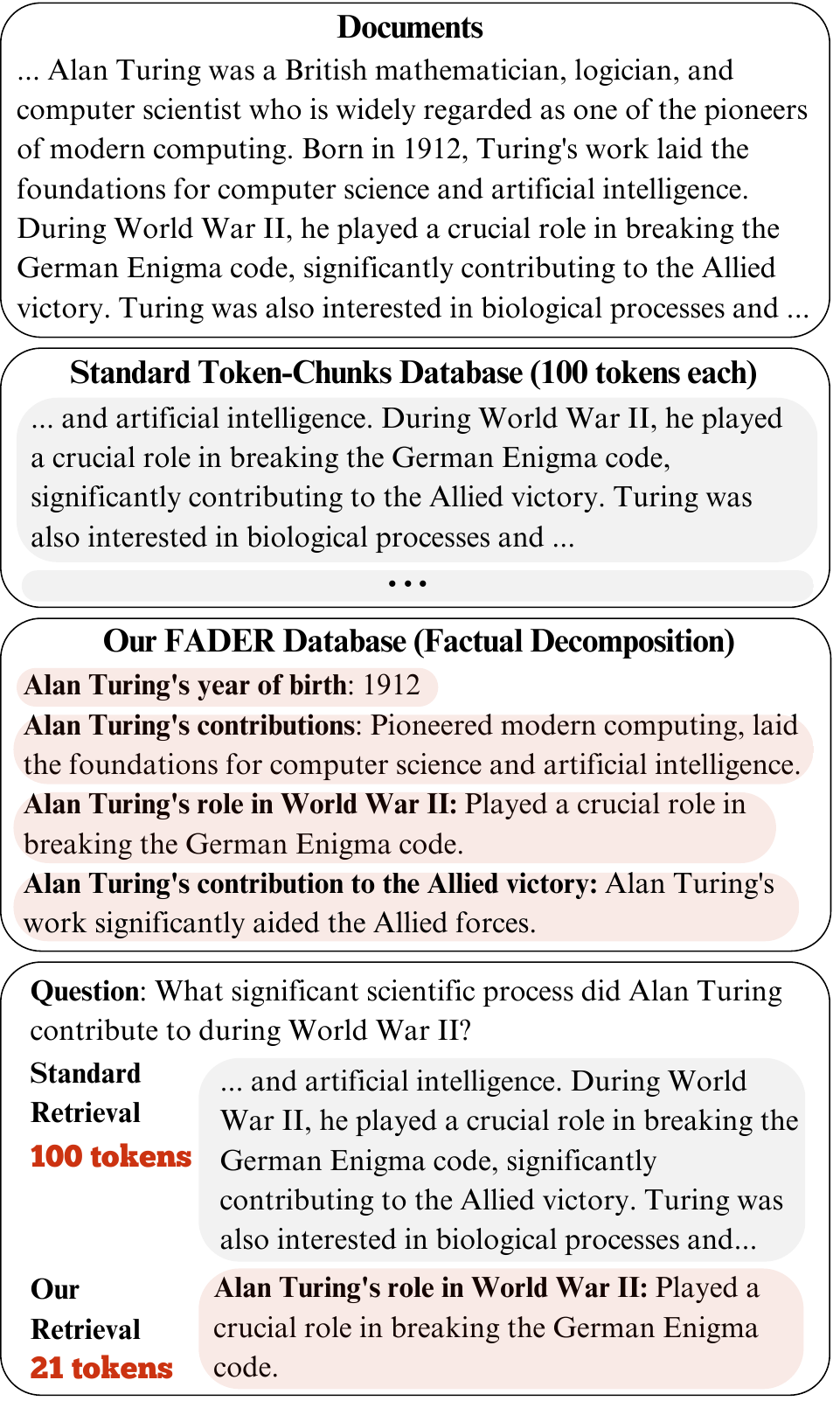}
    \caption{Example of datastore used for knowledge retrieval in our approach compared with typical fixed-size text chunks in RAG. We retrieve much shorter contexts.}
    \label{fig:datastore_example}
\end{figure}

Although large language models (LLMs) demonstrate remarkable capabilities across various tasks, their inability to continuously adapt to dynamic or domain-specific knowledge without parameter updates remains a substantial limitation.
To address this limitation, retrieval-augmented generation (RAG) supplements models with some external knowledge source
during inference \citep{lewis2020retrieval, pmlr-v162-borgeaud22a, ram2023context, li2024chunk}.

Typically these models treat the external source as a set of arbitrarily segmented blocks of raw text. However, there has also been interest
in using more structured external knowledge sources such as knowledge graphs
\citep{edge2024localglobalgraphrag,peng2024graph}, compressed documents
\citep{xu2024recomp} or document trees \citep{sarthi2024raptor}.  In each case one can identify a ``unit of retrieval'' where one retrieves some set of such units, such as a set of documents or a set of knowledge graph triples. Various candidates for units of retrieval, such as different forms of ``atomic facts,'' have been formulated  \citep{chen2023densex, jiang2024longrag,min-etal-2023-factscore, gunjal2024molecularfactsdesideratadecontextualization}.

In designing units of retrieval there is a tension between concise but brittle logical representations, such as knowledge graph triples, and highly expressive and nuanced, but verbose and unstructured, chunks of raw text. We propose an intermediate retrieval unit that we call an entity-description pair (EDP).  This is a pair of an ``entity''\footnote{Here we take a very liberal notion of ``entity'' not to be confused with the narrow notion of entity used in named entity recognition.} and some form of description of that entity.  For example
the entity might be ``Alan Turing's contributions'' and the factual description could be ``Pioneered modern computing, laid the foundation of computer science and engineering.'' Each EDP is a structured piece of information, like in structured databases, but also enjoys the flexibility of natural language. See Figure \ref{fig:datastore_example}.
We use a three-step language model prompting protocol to decompose a chunk of free text into a collection of EDPs and use the EDPs as the unit of retrieval in the resulting EDP knowledge base (KB).
Our approach, referred to as \textbf{FADER} (\textbf{F}actual \textbf{A}tomic \textbf{D}ecomposition for \textbf{E}fficient \textbf{R}etrieval), aims to reduce RAG inference overhead by pre-processing external corpora into atomic facts for efficient, high-precision retrieval.

Our main result is a demonstration that on various challenging question answering benchmarks EDP KB retrieval with FADER achieves better accuracy when the amount of retrieval (the number of retrieved tokens) is limited.
This can be phrased as improving the ``context-efficiency'' of RAG. We are also optimistic that our formulation of EDP KBs is a significant step
toward more structured yet expressive internal representations of knowledge.

\section{Related Work}
\label{sec: related_work}

\paragraph{Context-Efficient Retrieval}

As we will see in experiments, our approach achieves superior performance in \emph{context-efficient retrieval}, which we define as RAG methods aiming to reduce retrieved contexts for cost-effective LLM generations.
Previous related work involves various compression methods.
Some focus on vector-based compression, where models learn to compress long contexts into compact memory slots through end-to-end training \citep{ge2024incontext, cheng2024xragextremecontextcompression}. Others are text-based compression, which includes training rerankers \cite{pradeep2023rankvicuna}, applying extractive summarization \cite{xu2024recomp}, or training abstractive summarizers to compress the retrieved context \citep{xu2024recomp, jiang-etal-2024-longllmlingua}.
We also reduce retrieved contexts, but rather than compressing them \textit{post-retrieval}, we achieve context efficiency from the outset through improved knowledge representation \textit{pre-retrieval}.
Post-retrieval context compression methods still rely on token chunks as coarse units of knowledge for retrieval, whereas we structure the knowledge more efficiently with clear, well-defined representations that maintain high expressivity.

\paragraph{Knowledge Representation for Retrieval}

Most previous works directly segment source documents into equal-length text chunks, each containing hundreds of tokens \citep{lewis2020retrieval, ram2023context, pmlr-v162-borgeaud22a}. Recent research has explored alternative formats for knowledge representation, such as indexing source documents using knowledge graphs \citep{edge2024localglobalgraphrag}, hierarchical tree structures \cite{sarthi2024raptor}, or more relaxed versions of knowledge graphs \cite{liang2024empoweringlargelanguagemodels}.

Unlike these works, we don't rely on any explicit relational structures; our knowledge datastore consists of flat, semi-structured \textit{entity-description pairs}. 
The work most similar to ours is \citet{chen2023densex}, which decomposes each sentence in Wikipedia into individual propositions for retrieval.
However, we propose a novel method where speculated queries generated by the LM guide the fact extraction, and repeated samples of factual decompositions enhance our database. These techniques yield significant performance improvements by enabling more targeted information extraction and increased coverage of constructed facts, resulting in more relevant and concise data being incorporated during inference.\footnote{More general background of RAG is in \cref{sec:extra_related_work}.}

\section{Methodology}
\label{sec: method}

\begin{figure}[tb]
    \centering
    \includegraphics[width=1.0\linewidth]{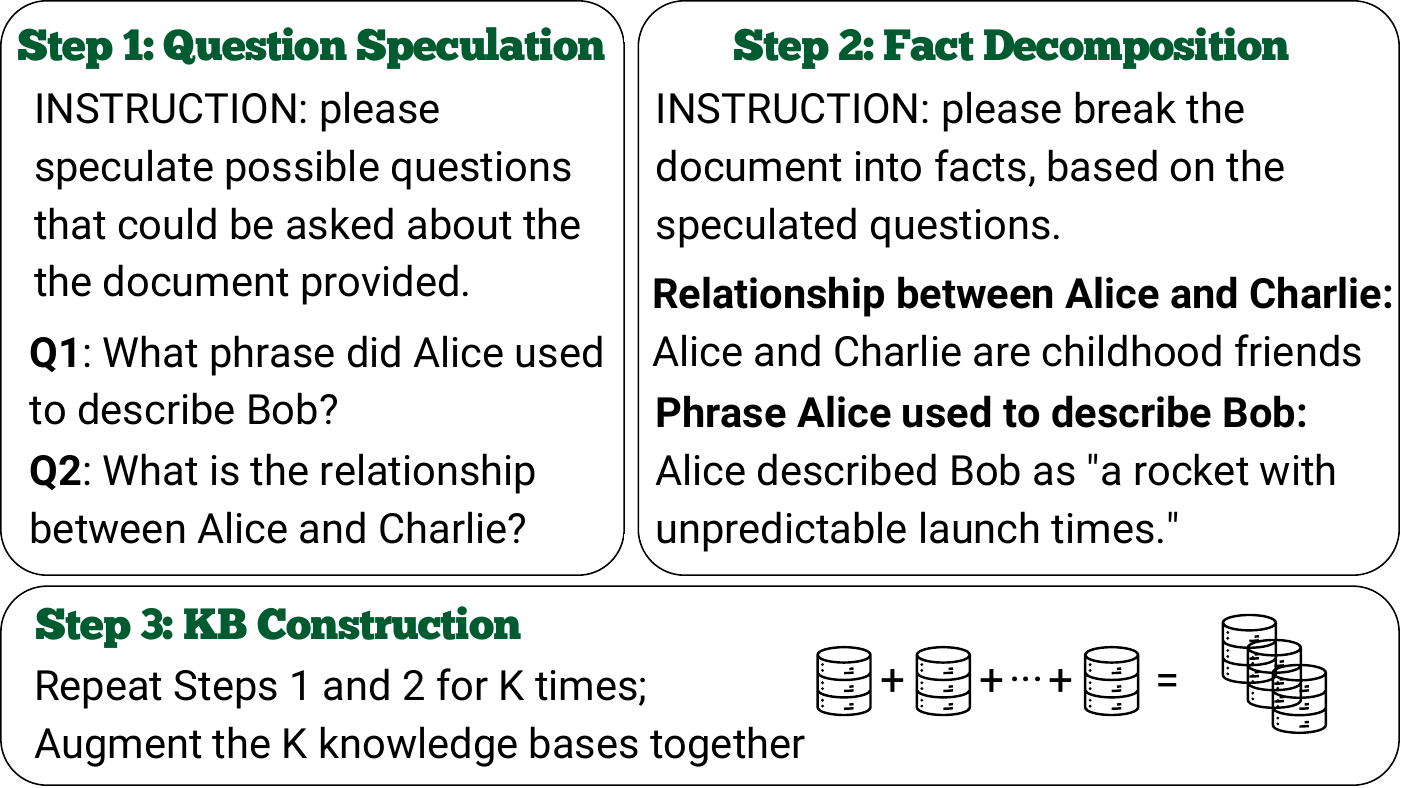}
    
    \caption{Overview of the \textsc{FADER} pipeline.}
    \label{fig:method}
\end{figure}

As shown in \cref{fig:method}, FADER consists of three main steps: (1) question speculation, (2) atomic fact extraction, and (3) knowledge base (KB) augmentation.
This approach allows to decompose long documents into concise, factual entity-description pairs (EDPs), building up a semi-structured KB for RAG, reducing retrieval overhead on the lengths of contexts to improve inference efficiency.

\subsection{Question Speculation}

Let $D$ represent a long document, which is split into $N$ equal-length chunks $D = \{D_1, D_2, \dots, D_N\}$, where each chunk $D_i$ contains approximately the same number of tokens. For each chunk $D_i$, we prompt a LM to speculate a set of possible questions $Q_i = \{q_{i1}, q_{i2}, \dots, q_{iJ}\}$, where $q_{ij}$ represents a potential question one might ask about the chunk $D_i$. This question speculation process helps direct the extraction of relevant knowledge and facilitates targeted information retrieval later.

Formally, given a document chunk \( D_i \), we define the question speculation process as: \( Q_i = \text{LM}_{\text{speculate}}(D_i) \), where \( \text{LM}_{\text{speculate}} \) denotes the question-speculation language model. The result is a set of speculative questions \( Q_i \) for each document chunk \( D_i \).

Prompts for $\text{LM}_{\text{speculate}}$ are shown in \cref{tab: ques_spec_prompts_v1} and \cref{tab: ques_spec_prompts_v2} in \cref{appendix: prompts}.

\subsection{Query-Guided Factual Decomposition}

Once we have the speculative questions $Q_i$, we feed both the set of questions $Q_i$ and the corresponding document chunk $D_i$ into the language model to extract relevant information that can be used to answer the questions. The goal is to retrieve concise, atomic facts that are highly specific and contextually relevant.

We prompt the language model to produce a set of EDPs for each chunk, where EDP is defined as a pair $k_{im}$ consisting of an entity $e_{im}$ and a fact $f_{im}$. The entity $e_{im}$ represents a key concept, while the fact $f_{im}$ encapsulates the essential information regarding $e_{im}$. Notably, the entity needs not be limited to a noun or an entry from a traditional knowledge graph; it can be a short noun phrase, sentence, or even a question.

The extraction process for a chunk \( D_i \) is as: \( K_i = \text{LM}_{\text{extract}}(D_i, Q_i) \), where \( K_i = \{k_{im} = (e_{im}, f_{im})\}_{m=1}^M \) is the set of EDPs for chunk \( D_i \). This method ensures that the extracted knowledge is both flexible and informative.
Note that each EDP $k_{im}$ does not have to correspond to a particular query $q_{ij}$ as $K_i$ are generated collectively with guidance from all $Q_i$, and the total number of EDPs $M$ could vary, regardless of the size of $Q_i$.
Prompts for $\text{LM}_{\text{extract}}$ are shown in \cref{tab: kb_prompts_v1} and \cref{tab: kb_prompts_v2} in \cref{appendix: prompts}.

\subsection{Sample Augmentation}

To further enrich the knowledge base, we apply a sampling-based approach that augments the fact extraction across multiple runs. By repeating the extraction process multiple times using the same prompt and leveraging the inherent randomness of the LM's outputs, we capture diverse sets of EDPs and prevent information gaps. We aggregate the knowledge extracted from these different runs to build a more comprehensive and robust KB.

Let \( S \) denote the number of sampling runs. For each document chunk \( D_i \), we repeat the extraction process, including both question speculation and EDP extraction, \( S \) times, yielding multiple KBs: \( K_{i}^{(1)}, K_{i}^{(2)}, \dots, K_{i}^{(S)} \). We then merge these KBs to form a final, augmented knowledge base \( K_i^{\textrm{final}} \) for chunk \( D_i \): \( K_i^{\textrm{final}} = \bigcup_{s=1}^{S} K_{i}^{(s)} \). The final knowledge base for the entire document \( D \) is then constructed by merging the augmented knowledge from all chunks: \( K^{\textrm{final}} = \bigcup_{i=1}^{N} K_i^{\textrm{final}} \). \( K^{\textrm{final}} \) provides a rich semi-structured knowledge repository for retrieval, where units are each EDP.

\section{Experiments}
\subsection{Setup}
\paragraph{Data}
Following prior work \cite{sarthi2024raptor}, we evaluate our method on three long-context QA datasets:
\begin{itemize}[leftmargin=*]
\item 
\textbf{NarrativeQA \cite{kovcisky2018narrativeqa}} consists of questions based on books and movie transcripts, requiring comprehension of entire stories. We report BLEU-4 \cite{BLEU}, ROUGE-L \cite{lin-2004-rouge}, and METEOR \cite{banerjee-lavie-2005-meteor} scores on the test set to measure the quality of generated answers, following previous work \cite{sarthi2024raptor}.
\item
\textbf{Qasper \cite{qasper-dataset}} includes questions from NLP research papers, focusing on detailed information extraction from full texts. Answers are categorized as Answerable/Unanswerable, Yes/No, Abstractive, and Extractive. We evaluate using the F1 metric on the test set, reflecting the overlap between predicted and reference answers, following previous work \cite{sarthi2024raptor}.
\item
\textbf{QuALITY \cite{pang-etal-2022-quality}} contains multiple-choice questions paired with context passages averaging around 5,000 tokens from various English articles (e.g., sci-fi, magazine articles, nonfiction). Since the test set is not public, we report accuracy on the validation set, measuring the proportion of correctly answered questions, following previous work \cite{sarthi2024raptor}.

\end{itemize}


\paragraph{Implementation}
We use BM25 \cite{bm25} as the retriever for both standard retrieval and our method, due to its effectiveness in prior studies. For our EDP-based knowledge base construction, we employ ChatGPT (gpt-4-2024-08-06) \cite{openai2024gpt4technicalreport}, which generates entity decomposition propositions efficiently. For question answering, we use Mixtral-8x7B-Instruct-v0.1 \cite{jiang2024mixtralexperts}, a state-of-the-art instruction-tuned language model suitable for downstream QA tasks.

\paragraph{Baselines}
Following RECOMP \cite{xu2024recomp}, we compare our approach to the following baselines to ensure a fair evaluation:
\begin{itemize}[leftmargin=*]
\item
\textbf{Standard Retrieval} applies BM25 on raw document chunks without any decomposition or summarization. 
\item
\textbf{Decomposition into Propositions \cite{chen2023densex}} uses ChatGPT to decompose documents into propositions, aiming to enhance retrieval by indexing finer-grained units.
\item
\textbf{Retrieve-then-Summarize}, as the core idea underlying RECOMP \citep{xu2024recomp}, utilizes off-the-shelf summarizers like T5-large \cite{raffel2023exploringlimitstransferlearning} and GPT-3.5 \cite{brown2020languagemodelsfewshotlearners} to condense retrieved documents before answering.\footnote{We do not use the learning-based extractive or abstractive compression methods in RECOMP because GPT-3.5-turbo---the teacher model from which they distill---consistently outperforms those variants (see Table 2 in \citet{xu2024recomp}). Therefore, we adopt GPT-3.5-turbo for abstractive summarization as our baseline.}

\end{itemize}

\begin{figure}[t]
    \centering
    \includegraphics[width=\linewidth]{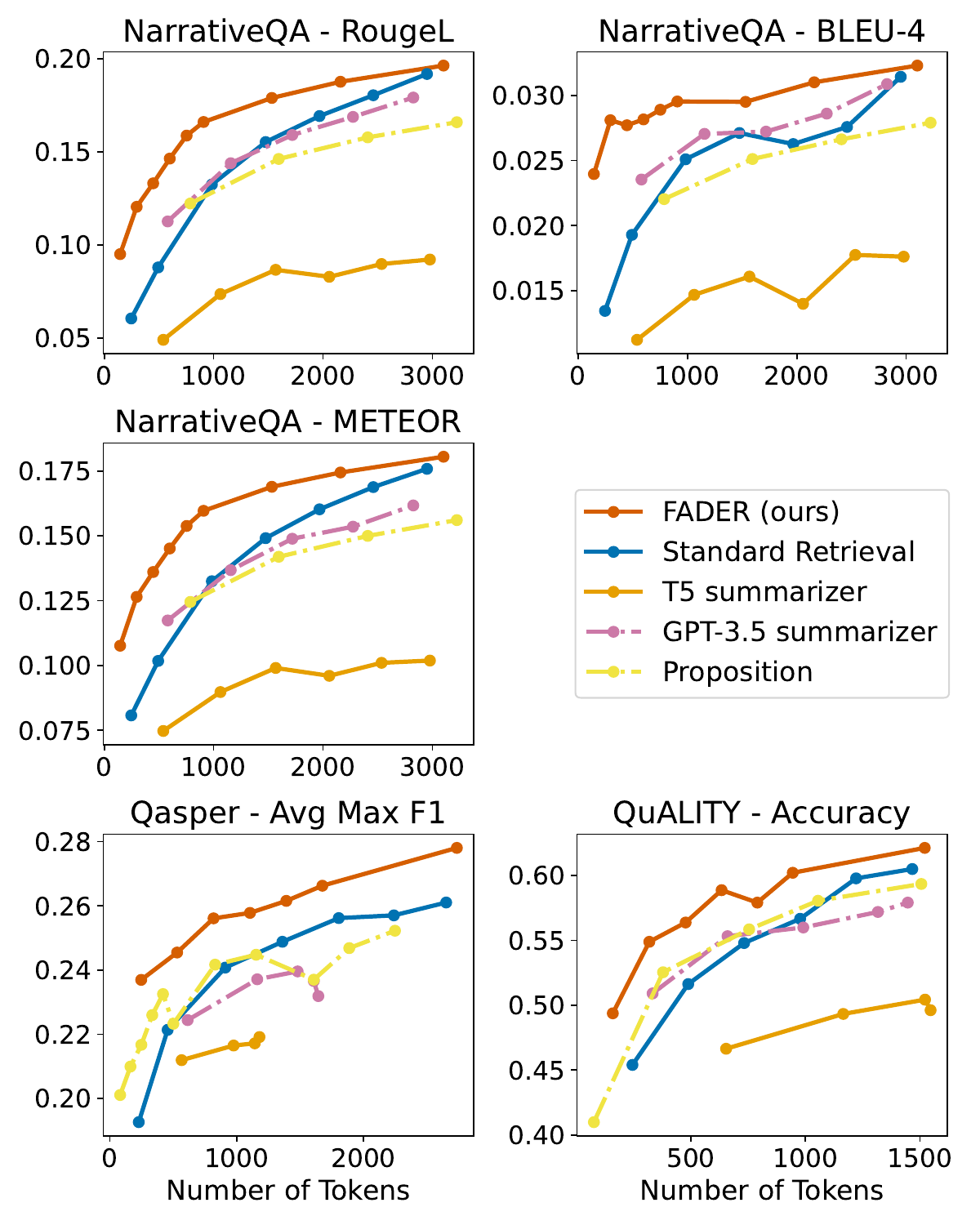}
    \caption{Results on NarrativeQA (top three plots), Qasper (bottom right), and quality (bottom left). The x-axis represents the number of tokens fixed in the retrieval context, and y-axis are different QA metrics used for each dataset.}
    \label{fig:all_KB_vs_baseline}
\end{figure}

\subsection{Context-Efficiency Evaluation}

Unlike most RAG methods that solely focus on downstream task performance like QA accuracy \citep{borgeaud2022improving, ram2023context} as evaluation, we propose to measure holistic RAG performance with an efficiency-aware metric through retrieval context budgets, which directly impacts LLM inference costs. This evaluation promotes retrieval methods that minimize LLM inference overhead, enhancing efficiency and potentially improving explainability by focusing on key information more effectively.

In particular, for each RAG method, we control the retrieval context budget $b$ provided to LLM for inference, measured by the number of retrieved tokens, and compute the downstream task performance metric $s$ for each budget $b$. Instead of focusing on a single best $s$, we examine the downstream task performance over the full range of context sizes, collecting a wide range of $(b, s)$ pairs. They are represented as a \textit{context-efficiency curve} (see example in \cref{fig:all_KB_vs_baseline}), capturing the tradeoff between the downstream task performance and LLM inference cost imposed by retrieval.
The goal is to maximize the Pareto frontier of the curve. Retrieval methods with higher context-efficiency curves indicate more efficiency with comparable task accuracy.\footnote{Previous studies have conducted similar analyses to our context-efficiency curve \citep{chen2023densex,  jiang2024longrag, yoon-etal-2024-compact}, but mostly as supportive metric for artifacts such as long contexts and their compression. In contrast, we propose the context-efficiency curve as a primary efficiency-aware performance measure for RAG.}

\subsection{Main Results}
\label{sec: result}

\cref{fig:all_KB_vs_baseline} shows results on NarrativeQA, Qasper, and QuALITY. We see that our method, FADER, consistently outperforms all baselines when the number of tokens in the context is kept at \textit{all} different levels, shown with our context-efficiency curve above all others. Our method performs especially well in the short-context regime when the retrieved tokens are very limited. 
This can effectively reduce LLM inference cost with more efficient usage of context in RAG.
We also find that decomposing sentences into propositions \cite{chen2023densex} does not generalize well to domains with lower fact density, such as novels or scientific papers. Some qualitative examples of retrieved documents for each method are provided in \cref{sec: retrieved_examples}.

\subsection{Analysis}
\label{sec: method_ablations}

Here we ablate each component of our method (\cref{sec: method}) on their contribution to the overall performance.

\begin{figure}[tb]
    \centering
    \includegraphics[width=0.95\linewidth]{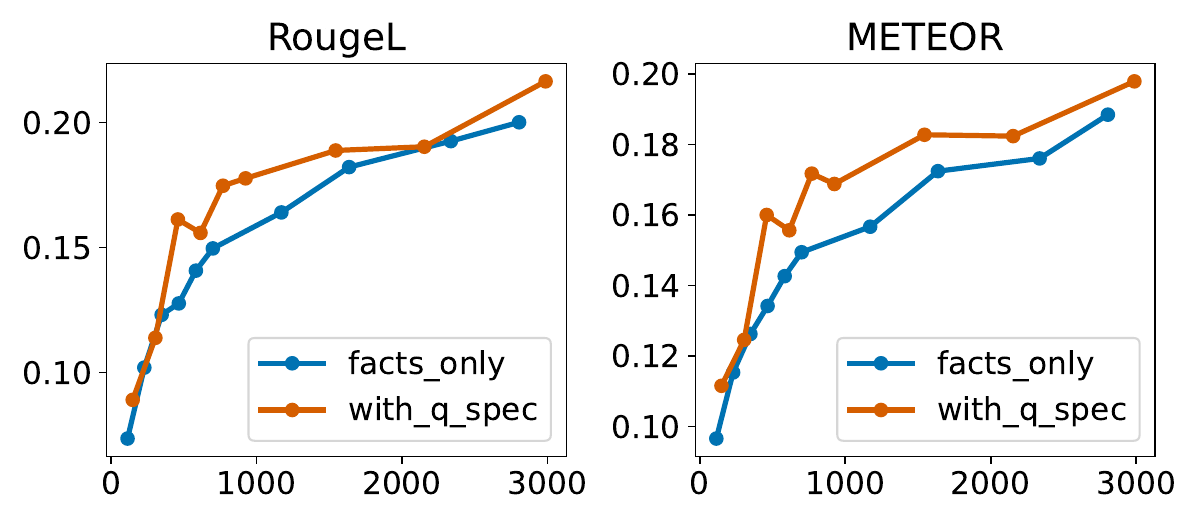}
    
    \caption{Comparison of performance (y-axis) vs. number of retrieved tokens (x-axis) between Fact-Only KB construction and Question-speculated KB construction on a subset of NarrativeQA's validation set.}
    \label{fig:narrativeqa_q_spec_ablations}
\end{figure}

\begin{figure}[tb]
  \centering
    \includegraphics[width=0.95\linewidth]{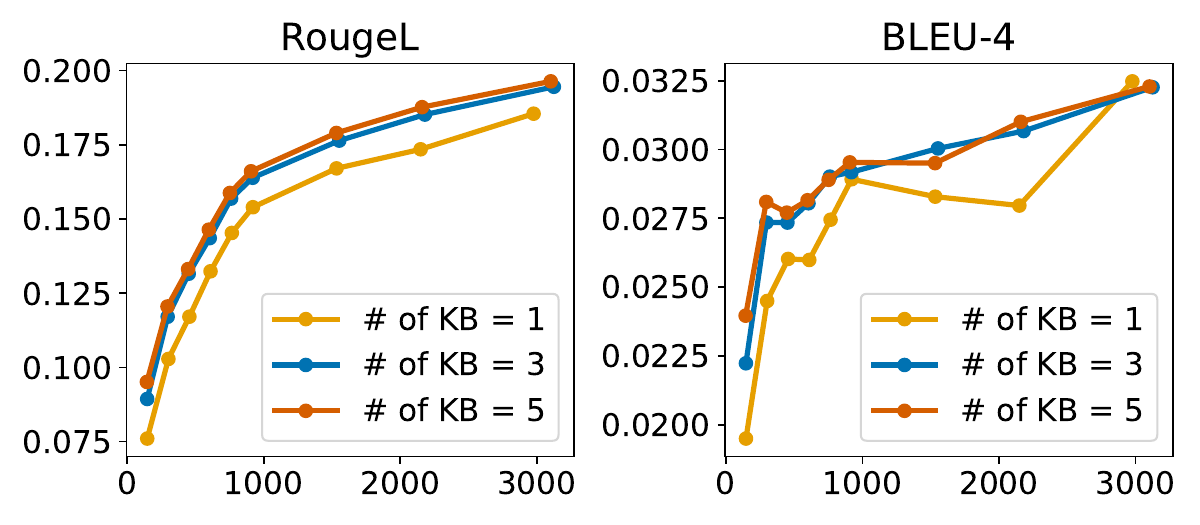}
    \caption{Performance (y-axis) vs. number of retrieved tokens (x-axis) on NarrativeQA for the number of resampled KBs equal to 1, 3, and 5.}
    \label{fig:narrativeqa_KB_num_KBs}
\end{figure}

\paragraph{Why Question Speculation?}

 Despite extensive prompt tuning, providing speculated queries to LM when generating the KB consistently yields better performance compared to letting the LM extract facts without guidance (see \cref{fig:narrativeqa_q_spec_ablations}). To investigate this, we randomly select 20 stories (617 associated queries) from NarrativeQA and compute the similarities between the speculated questions and the real queries. Surprisingly, using similarity thresholding heuristics and manual inspection, we find that 11.18\% of the speculated questions closely align with or rephrase the real queries, and 53.97\% focus on the same topic (for more details, see \cref{sec: q_spec_question_sim}). This significant overlap aligns with previous research showing that LLMs are effective at generating synthetic queries \cite{wu2024llm}. In fact, these speculated questions function like a chain-of-thought process \cite{chain-of-thought}, allowing the LM to gather relevant information before answering the query.

\paragraph{Why KB Augmentation?}

We observe that the questions speculated and facts extracted vary between different runs due to the LM's inherent stochasticity. \cref{fig:narrativeqa_KB_num_KBs} shows that augmenting KBs improves performance, indicating that the sampling process effectively captures a more diverse range of meaningful knowledge pairs. Full results for NarrativeQA, Qasper, and QuALITY are in \cref{sec:kb_augmentation_abalation}.

\subsection{Quality Checks on Speculative Questions and EDPs}
\label{sec:quality_checks}

While our method demonstrates strong performance, we carefully evaluate the quality of the speculative questions that guide fact extraction and the generated EDPs.

\paragraph{Automatic Evaluation of Speculative Questions.}
As a proxy for assessing the quality of the generated questions, we measure their similarity to real queries in the validation set of the corresponding dataset (using 20\% of that set for this evaluation). Following standard practice, we employ an embedding-based similarity approach using the all-MiniLM-L6-v2 model from Hugging Face's sentence-transformers.\footnote{Available at \url{https://huggingface.co/sentence-transformers/all-MiniLM-L6-v2}.}
The higher the average similarity scores, the more closely the speculative questions resemble real queries, which is desirable. Examples of some of the highest-similarity pairs can be found in \cref{sec:appendix_examples} (\cref{tab:sim_examples}).

\paragraph{Manual Evaluation of EDPs.}
To ensure consistency and accuracy of EDPs, we also conduct a thorough manual evaluation. Specifically, we randomly select 200 examples from our generated datastore for each dataset (NarrativeQA, Qasper, QuALITY). A team of three reviewers independently assessed the quality, coherence, and correctness of the EDPs. We did not identify any contradictions or significant issues in these sampled EDPs.

\section{Conclusion}
\label{sec: conclusion}

We propose FADER, a retrieval-augmentation pipeline that decomposes external corpora into atomic facts as units of retrieval that focuses on efficient and high-precision inference.
We demonstrate that on various challenging question answering benchmarks, FADER achieves better accuracy when the amount of retrieval (the number of retrieved tokens) is limited.
This improves the \textit{context-efficiency} of RAG, which has real cost-effective implications for LLM inference. We are also optimistic that our formulation of EDP KBs is a significant step
toward more structured yet expressive internal representations of knowledge, and we encourage future research to build upon and expand this approach.

\section*{Limitations}

While our approach demonstrates improved context-efficiency in retrieval-augmented generation for question answering tasks, several limitations warrant discussion. First, our method relies heavily on the performance of large language models for both question speculation and factual decomposition. Any biases or errors inherent in these models could propagate through the process, potentially affecting the quality and reliability of the extracted entity-description pairs.

Second, the stochastic nature of our sampling-based augmentation introduces variability in the generated knowledge bases. Although multiple samples help capture a broader range of information, this approach may lead to inconsistencies across different runs. Further research is needed to assess the stability and reproducibility of the results when applying our method in diverse settings.

In summary, while our method enhances context-efficiency, it remains vulnerable to inherent LLM biases and sampling-induced variability. Addressing these issues is crucial for improving the reliability and consistency of our approach in various applications.
\bibliography{reference}

\appendix

\clearpage
\section*{Appendix}

\section{Background on RAG}
\label{sec:extra_related_work}

\paragraph{Retrieval-Augmented Generation}

Retrieval-augmented generation~\citep{lewis2020retrieval} (RAG) is the process of dynamically adding additional information at inference time through a similarity search process in order to improve generation quality. It is typically used in domains where it may be difficult for the language model to rely on parametric knowledge alone, for example long-tail question answering, or for current events past the training data cutoff date. The simplest form of RAG is to add text related to the query directly to the input~\citep{ram2023context}. There are also vector-based variants, for example injecting information at deeper layers of the network~\citep{borgeaud2022improving, wu2022memorizing, bertsch2023unlimiformer}, or interpolating with a nearest neighbor generation~\citep{khandelwal2019generalization}. Some works tune with retrieval-augmentation~\citep{guu2020retrieval}, or to induce retrieval behavior~\citep{asai2024self, li2024chunk}. Though retrieval-augmentation is generally quite beneficial, language models can be distracted depending on the order~\citep{liu2024lost} or content~\citep{yoran2023making} of the data retrieved.

\section{Experiment Setup}

\subsection{Datasets}
\label{sec:detaset_detail}

\begin{itemize}
    \item \textbf{NarrativeQA} \cite{kovcisky2018narrativeqa} is a dataset containing 1,572 documents, including books and movie transcripts. It requires answering questions based on the full text of these narratives. The task tests the model’s ability to comprehend entire stories, with performance measured using BLEU (B-1, B-4), ROUGE (R-L), and METEOR metrics. We report BLEU-4, ROUGE-L and METEOR on the entire test set.
    
    \item \textbf{QASPER} \cite{qasper-dataset} consists of 5,049 questions drawn from 1,585 NLP papers, with answers categorized as Answerable/Unanswerable, Yes/No, Abstractive, and Extractive. The questions focus on extracting detailed information embedded within the full text of the papers. Accuracy is evaluated using the F1 metric, reported on the entire test set.
    
    \item \textbf{QuALITY} \cite{pang-etal-2022-quality} contains multiple-choice questions, each paired with context passages averaging around 5,000 tokens.  Since the QuALITY test set is not public, accuracy is reported on the validation set.
\end{itemize}

\subsection{Details on Setup}
\label{sec:setup_detail}
For the standard retrieval baseline, we experiment with different token counts within a chunk (see \cref{sec:chunk_len_ablation}) and select the best-performing one as the final baseline. In all experiments, we follow \citet{sarthi2024raptor}, using \textsc{cl100k\_base} from Tiktoken as the tokenizer to split source documents into chunks and compute final token usage. We use BM25 as the retriever, ChatGPT (gpt-4o-2024-08-06) for our EDP-based KB construction, and Mixtral-8x7B-Instruct-v0.1 for question answering.

\subsection{Best Chunk Length for Standard Retrieval Baseline}
\label{sec:chunk_len_ablation}

\begin{figure*}[h!]
    \centering
    \includegraphics[width=1\linewidth]{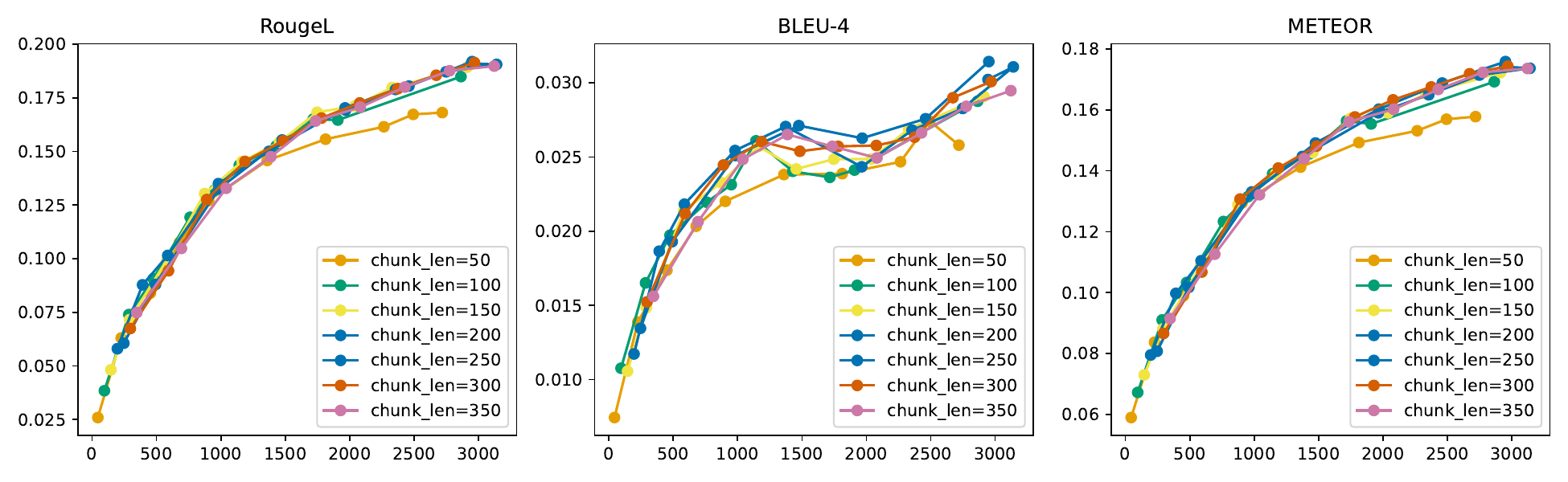}
    \caption{Metrics vs Number of Tokens for Different Chunk Lengths on NarrativeQA}
    \label{fig:narrativeqa_baseline_chunklen}
\end{figure*}

\begin{figure}[h!]
    \centering
    \includegraphics[width=0.8\linewidth]{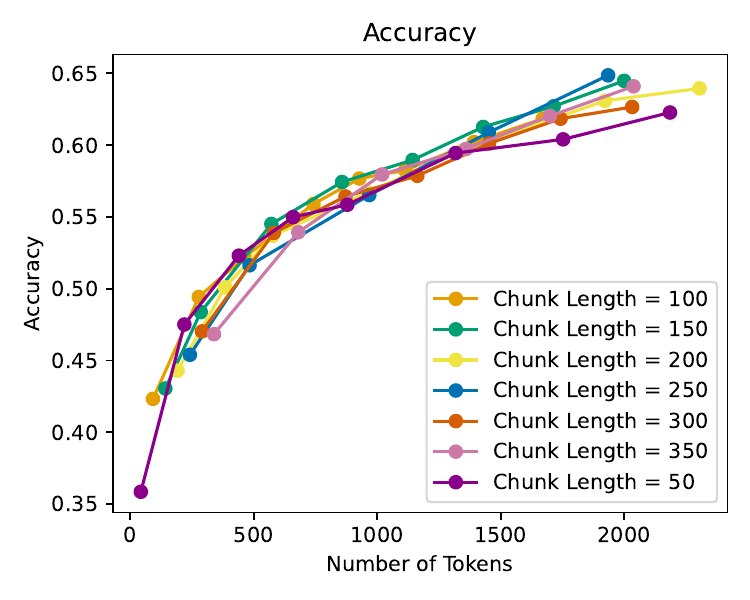}
    \caption{Accuracy vs number of retrieved tokens for different chunk lengths as retrieval units in the standard RAG approach on QuALITY dataset.}
    \label{fig:quality_chunklen}
\end{figure}

\begin{figure}[h!]
    \centering
    \includegraphics[width=0.8\linewidth]{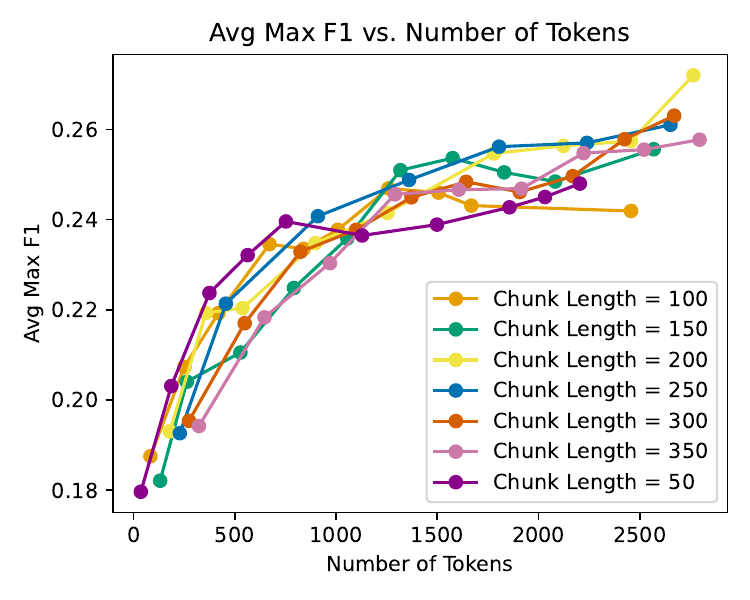}
    \caption{Avg Max F1 vs number of retrieved tokens for fifferent chunk lengths as retrieval units in the standard RAG approach on Qasper dataset.}
    \label{fig:qasper_chunklen}
\end{figure}

We perform comprehensive ablation studies to find the optimal chunk length for each retrieved document (see \cref{fig:narrativeqa_baseline_chunklen}, \cref{fig:qasper_chunklen}, \cref{fig:quality_chunklen}). We test chunk lengths of 50, 100, 150, 200, 250, 300, and 350 tokens, ensuring sentence boundaries are respected when chunking the book into fixed-size documents. For each chunk length, we select 5-10 different numbers of documents. We find that a chunk length of 250 tokens achieves the best performance on NarrativeQA, Qasper, and QuALITY, and we use this as the naive retrieval baseline reported in the main text.

\section{Qualitative Examples of Retrieved Documents }
\label{sec: retrieved_examples}

We provide datastore examples that are retrieved when answering a question from NarrativeQA. \cref{tab: retrieved_examples_ours_standard} shows our retrieval compared to the standard retrieval, and \cref{tab: retrieved_examples_proposition} shows the retrieval following \citet{chen2023densex}'s proposition method. We find that our retrieval leads to the best final answer, while the other two approaches struggle to retrieve the correct information from their datastores. The standard baseline fails to find the relevant chunk from the book, and the proposition baseline decomposes all human dialogue into even smaller units, which makes the information more scattered and harms retrieval.

\begin{table*}[h!]
\centering
\small
\begin{tabular}{@{}p{0.95\textwidth}}
\toprule
\textbf{Question}: What does Mark urge his listeners to do? \\
\textcolor{purple}{\textbf{Ground Truth}: to do something about their problems instead of committing suicide} \\
\midrule
\textbf{Our retrieval}: 

- "how does happy harry hardon describe his listeners?: happy harry hardon refers to his listeners as 'horny'.", 

- "what actions does happy harry hardon encourage his listeners to take when contacting him?: happy harry hardon encourages his listeners to send him their most pathetic moments, as long as they are real, and wants details like size, shape, feel, and smell.", 

- "who is happy harry hardon and what role does he play?: happy harry hardon is a radio host, engaging with his listeners over the air, and helping chris discuss his experiences and feelings.", 

- "what contest does happy harry hardon mention during his show and how does he engage his listeners?: happy harry hardon runs a contest on the best way to put parents out of their misery and challenges listeners to amaze and discuss the sensational content of his show.", 

- "what concern does mark have about the effects of his message?: mark feels ill about the situation and perceives it as chaotic, tangled with his life", 

- "what does mark express about his feelings in the conversation with his parents?: he finds it difficult to talk to girls in his new place and feels different.", 

- "what reaction does mark have to the police car pulling up?: mark is not worried and thinks the police are just dropping in on his neighbor.", 

- "what internal conflict does mark express after his interaction with nora?: mark expresses that the whole thing is making him ill and that it's his life that nora is messing with.", 

- \textcolor{blue}{"what critical message does mark want to convey to his audience about taking control of their lives?: mark conveys that it's their life, they should take charge of it, and they should fill the air and speak out."}, 

- "english class question: jan emerson asks mark to share his feelings about what he wrote."

\textcolor{purple}{\textbf{Our answer}: take charge of their lives.} \\
\midrule

\textbf{Standard retrieval baseline}: 

"happy harry hardon - you interview a student and then you rat on her, you betray her  trust, isn't that right sir <deaver hangs up> well as you can see, these guys are played  out  society is mutating so rapidly that anyone over the age of twenty has really no  idea  err alright, back down to business  \"i share a room with my older brother and  nearly every night after he turns off his light he come over to my bed and gives me a few  arm nookies and stuff and then makes me scratch his back and other refinements\" it's  about time we had some refinements on this show  \"then sooner or later he gets worked  up and further a do he rubs his thing and makes me watch \" signed \"i'm just screwed up\"  well first of all you're not screwed up, your an unscrewed up reaction to a screwed up  situation  feeling screwed up at a screwed up time, in a screwed up place does not make  you necessarily screwed up, if you catch my drift  well as you know dear listeners if you  enclose your number a reply is guaranteed  <rings miss screwed up>", 

"creswood - it's the trouble makers, you can't run a top school with trouble makers in the  mix brian - okay, so what exactly is a trouble maker creswood - someone who has no interest in education brian - oh c'mon that includes every teenager i know creswood - can't you understand that nothing is more important than a good education brian -  except for the basic right to it creswood - the point is i have the highest s a t  scores in the state brian - yeah but how creswood - i stand by my record <the school field> shep sheppard - mr  watts, shep sheppard channel six news here watts - good evening shep sheppard - how does washington intend to deal with this situation watts - we at the f c c  feel that democracy is about protecting the rights of the ordinary  citizen  un-regulated radio would result in programming of the lowest common  denominator, the rule of the mob  <watts looks round to see one of his f c c  vans has  picked up some graffiti> this is vandalism, not free expression <everyone gathers outside the school to listen to the happy harry hardon show>"

\textcolor{purple}{\textbf{Standard retrieval's answer}: mark does not urge his listeners to do anything}\\

\bottomrule
\end{tabular}
\caption{Example of answering a question from NarrativeQA: In our method, the blue highlight represents the evidence most likely to contribute to the final answer.}
\label{tab: retrieved_examples_ours_standard}
\end{table*}

\begin{table*}[h!]
\centering
\small
\begin{tabular}{@{}p{0.95\textwidth}}
\toprule
\textbf{Question}: What does Mark urge his listeners to do? \\
\textcolor{purple}{\textbf{Ground Truth}: to do something about their problems instead of committing suicide} \\
\midrule

\textbf{Proposition Baseline \cite{chen2023densex}}: 

"creswood states, \"so what does this prove, not everyone goes to college.\"", 

"mark goes to collect his post.", 

"the speaker asserts, \"happy harry hardon will go to any language to keep his three listeners glued with huwy bluwy to their radios.\"", 

"happy harry hardon asks, \"are you willing to tell my listeners what you told me here in this letter?\"", 

"happy harry hardon invites listeners to share their most real moments.", 

"nora finds mark burning his happy harry hardon letters.", 

"mark adds, \"i know exactly what it means.\"", 

"mark protests, \"i swear, what are you doing?\"", 

"mrs. kaiser invites malcolm to join his parents downstairs.", 

"happy harry hardon states that listeners are interested in the decision to expel cheryl bates.", 

"mark explains that it is his mom's jeep and that she kind of loaned it to him.", 

"chris expresses, \"i didn't know what to do.\"", 

"mark says, \"no it's outside,\" and shows nora his converted radio jeep.",

"happy harry hardon asks david deaver to explain his work.", 

"mark collects his post from the postal center, exits, and starts to read the eat me beat me lady's letter.", 

"donald shakes his head in disgust.", 

"chris asks, \"so what are we going to do about this?\"", 

"happy harry hardon suspects a lie if miss screwed up does not remember or tell the truth.", 

"happy harry hardon concludes, \"but you know what you have to do.\"", 

"happy harry hardon lists, \"you have parents, teachers telling you what to do.\"", 

"nora pulls mark into the clayroom and reassures, \"it's cool, it's safe. guess what i heard?\"", 

"back outside the lockers, doug asks donald, \"so what did they do to you?\"", 

"nora questions, \"mark what is with you?\"", 

"malcolm's mother, mrs. kaiser, asks malcolm about his homework.", 

"happy harry hardon continues, \"you have movies, magazines, and tv telling you what to do.\"", 

"happy harry hardon questions what david deaver says to young people about the world's trustworthiness.", 

"detective denny, holding up his badge, implies that the postal clerk can give the information to him.", 

"mark asks, \"close to what?\"", "malcolm tells mrs. kaiser that he has finished his homework.", 

"happy harry hardon notes, \"now they've all run home to tune in and listen to what they've all been talking about.\"", 

"mark comments, \"yeah, back to you.\"", 

"happy harry hardon addresses his audience as \"all my horny listeners.\"", 

"marla hunter asks brian hunter, \"have you noticed his behaviour lately?\"", 

"brian questions, \"okay, so what exactly is a troublemaker?\"", 

"nora points out, \"f.c.c. you know what that means.\"", 

"happy harry hardon asks, \"so what did you do?\"", 

"happy harry hardon prompts, \"so tell us what happened.\"", 

"mark adds, \"i can't talk to them!\"", 

"mark mentions having something to show nora.", 

"mark comments to nora, \"you're so different.\"", 

"mark clarifies, \"i can't talk to you.\"", "nora greets, \"hi! what are you doing? you having fun?\"", 

"brian asks, \"loretta what the hell is going on here?\"", 

"cheryl asks, \"can you tell me what this is about?\"", 

"creswood asserts, \"nonsense, she doesn't know what she's talking about.\"", 

"happy harry hardon claims, \"happy harry just happens to have in his very hands a copy of a memo written by mr.\"", 

"mark asserts, \"i can't talk to you people.\"", 

"mark declares, \"steal it, it belongs to you.\"", "happy harry hardon acknowledges \"all of my horny listeners would love it if i would call up the eat me beat me lady.\"", 

"jan reveals, \"last night one of our students, malcolm kaiser, took his own life.\""

\textcolor{purple}{\textbf{Proposition Baseline's answer}: Mark does not urge his listeners to do anything. No specific action is mentioned.}\\

\bottomrule
\end{tabular}
\caption{Example for answering one question from NarrativeQA.}
\label{tab: retrieved_examples_proposition}
\end{table*}

\begin{table*}[h]
    \centering
    \small
    \begin{tabular}{p{4.5cm} p{8cm} c}
        \toprule
        \textbf{Real Question} & \textbf{Speculated Question} & \textbf{Similarity} \\
        \midrule
        \multicolumn{3}{l}{\textbf{Closely Related / Rephrase of the Question (Similarity $\geq$ 0.85)}} \\
        \midrule
        Why does Helen return to Grassdale? & Why does Helen eventually return to Grassdale alone? & 0.9637 \\
        What name does Klaatu use at the boarding house? & Where does Klaatu come from before entering the boarding house? & 0.9013 \\
        What object did Tom find in Klaatu's room? & What does Tom find on the floor of Klaatu's room? & 0.8852 \\
        How does Data finally defeat the Borgs? & What actions does Data take to thwart the Borg's attempts? & 0.8640 \\
        What gift did the Borg Queen offer Data? & What does the Borg Queen want from Data? & 0.8614 \\
        \midrule
        \multicolumn{3}{l}{\textbf{Questions on the Same Topic (Similarity 0.7 - 0.85)}} \\
        \midrule
        \multirow{6}{=}{What did Klaatu say would happen if his message was ignored by Earth's people?} & What does Klaatu want to discuss with representatives from Earth? & 0.7529 \\
        & What is Klaatu's demeanor when he discusses the stakes for Earth's future if his message is not heeded? & 0.7783 \\
        & How does Klaatu react to the replies from world leaders regarding the meeting? & 0.7284 \\
        & What alternative does Klaatu say Earth would face if his proposals are rejected? & 0.7517 \\
        & What message does Klaatu ask to be delivered and to whom? & 0.7136 \\
        & What ultimatum is being given to the audience in Klaatu's message? & 0.7272 \\
        \cmidrule{1-3}
        \multirow{2}{=}{Who did Bobby suggest was the greatest living person?} & How does Bobby respond to Klaatu's question about the greatest man in America? & 0.7343 \\
        & Who does Bobby identify as the greatest scientist in the world? & 0.7316 \\
        \bottomrule
    \end{tabular}
    \caption{Examples of Speculated Questions and Their Similarity to Real Questions}
    \label{tab:question_similarity}
\end{table*}
\newpage

\section{Ablations on Question Speculation}
\label{sec: q_spec_question_sim}

\cref{tab:question_similarity} shows the similarity between real queries and speculative queries in a subset of NarrativeQA. The similarity is measured by computing the similarity between embeddings encoded with the all-MiniLM-L6-v2 model from Hugging Face's sentence-transformers. We examined 617 questions and found that 11.18\% of the speculated questions closely align with or rephrase the real queries, while 53.97\% focus on the same topic.

\section{Ablations on KB Augmentation}
\label{sec:kb_augmentation_abalation}

\begin{figure*}[t]
    \centering
    \includegraphics[width=1\linewidth]{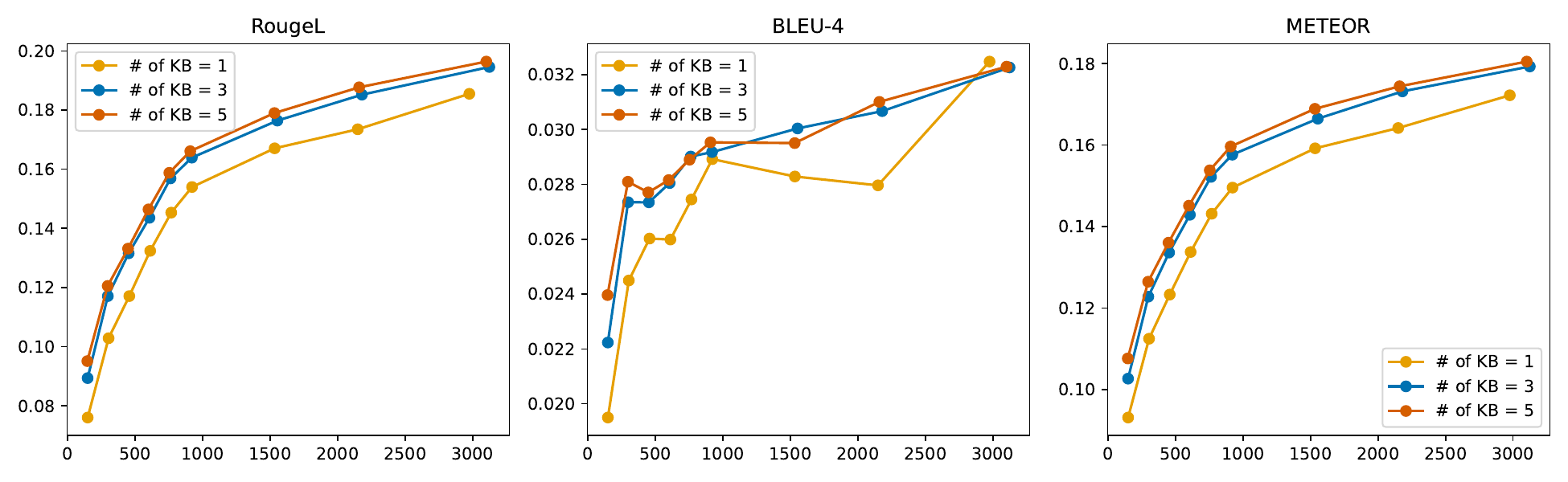}
    \caption{Results on different number of KBs on NarrativeQA.}
    \label{fig:narrativeqa_KB_ds_ablation}
\end{figure*}

\begin{figure}[h!]
    \centering
    \includegraphics[width=1\linewidth]{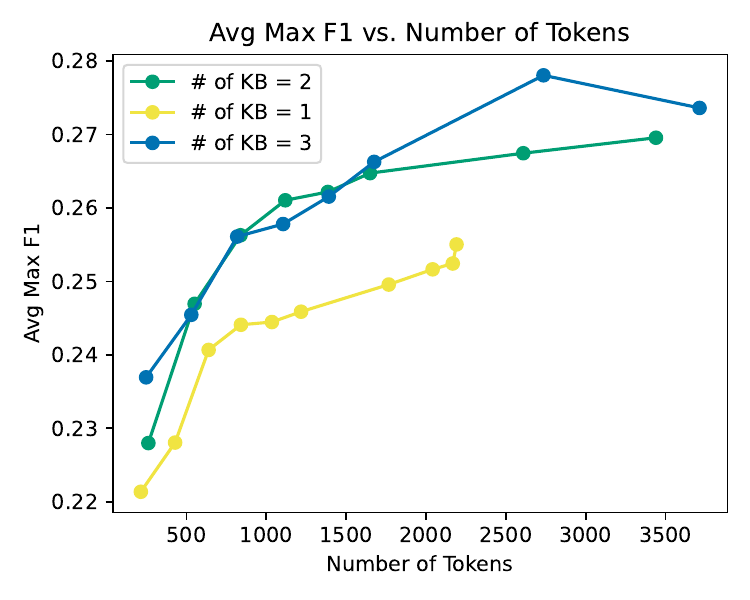}
    \caption{Results on different number of KBs on Qasper.}
    \label{fig:qasper_KB_ds_ablation}
\end{figure}

\begin{figure}[h!]
    \centering
    \includegraphics[width=1\linewidth]{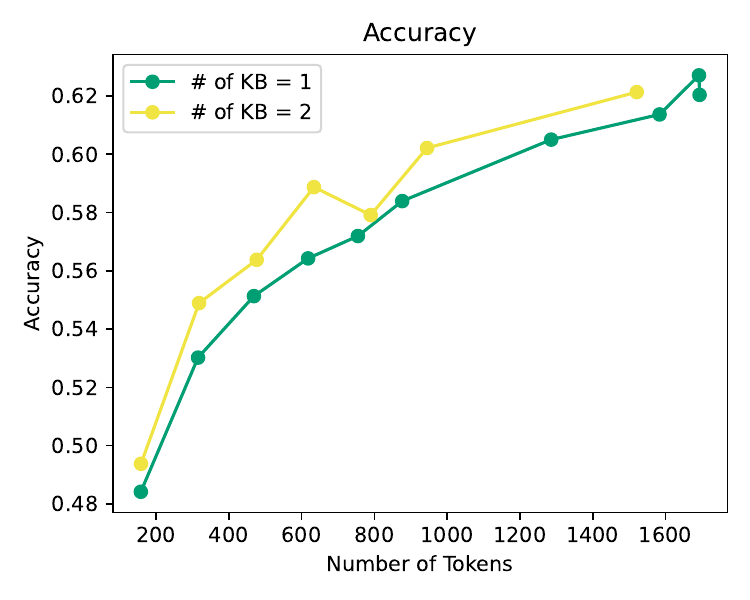}
    \caption{Results on different number of KBs on QuALITY.}
    \label{fig:quality_KB_ds_ablation}
\end{figure}

\cref{fig:narrativeqa_KB_ds_ablation}, \cref{fig:qasper_KB_ds_ablation} and \cref{fig:quality_KB_ds_ablation} 
 show the effect of different numbers of KBs in NarrativeQA, Qasper, and QuALITY.

\section{Additional Examples of Generated Questions}
\label{sec:appendix_examples}

Here, we provide additional examples of speculative questions and their real-query counterparts from NarrativeQA. \cref{tab:sim_examples} lists some of the highest-similarity pairs according to the all-MiniLM-L6-v2 model. These examples show that speculative questions are semantically aligned with real queries, which helps guide the LM to extract relevant facts without exact repetition.

\begin{table*}[h!]
    \centering
    \small
    \begin{tabular}{p{0.4\linewidth} p{0.4\linewidth} c}
    \toprule
    \textbf{Real Question} & \textbf{Speculative Question} & \textbf{Similarity}\\
    \midrule
    How does Liza get a black eye? & What causes Liza’s black eye? & 0.9264 \\
    What does Dr.\ Varava reveal about Esther? & What does Dr.\ Varava reveal to Kate about Esther? & 0.9189 \\
    What is Mr.\ Roundhay's profession? & What is Mr.\ Roundhay’s occupation and hobby? & 0.9327 \\
    \bottomrule
    \end{tabular}
        \caption{Highest-similarity speculative questions vs.\ real questions from the NarrativeQA validation set.}
    \label{tab:sim_examples}
\end{table*}

\section{Prompts}
\label{appendix: prompts}

We detail all the prompts used in our method and baselines. For our method, prompts for question speculation are shown in \cref{tab: ques_spec_prompts_v1} and \cref{tab: ques_spec_prompts_v2}. Prompts for EDP KB construction are shown in \cref{tab: kb_prompts_v1} and \cref{tab: kb_prompts_v2}. Prompts for question answering are shown in \cref{tab: qa_prompts_v1}, \cref{tab: qa_prompts_v2}, \cref{tab: qa_prompts_v3}, and \cref{tab: qa_prompts_v4}. Note that for NarrativeQA, we use a two-step prompting approach to obtain the final answer: first, perform regular question answering based on the query and retrieved documents; second, compress the answer to make it more concise. This is because answers in NarrativeQA are typically just a few words, but Mixtral tends to generate lengthy responses regardless of prompt adjustments, prompting us to adopt a two-step process.

\begin{table*}[h!]
\centering
\small
\begin{tabular}{@{}p{0.9\textwidth}}
\toprule
\textbf{NarrativeQA \& QuALITY (Question Speculation)} \\
\midrule
      \textbf{System}: You are a highly attentive assistant focused on generating specific and concise questions about the narrative elements of a text. Your goal is to produce clear and direct questions that help a reader deeply understand the concrete aspects of the story.
      \newline
      
    \textbf{User}: Task: Generate Specific, Concrete, and Contextual Narrative Questions
      \newline
      
      **Objective**: Given a section of text from the book, generate a set of specific, concise, and detailed questions that are directly related to the narrative elements—such as characters, actions, events, settings, and their historical or cultural significance. If the text contains irrelevant information like publisher details, web content, or other non-narrative elements, do not generate questions and instead return 'no questions extracted.'
      \newline
      
      **Instructions**:
      
      1. **Read the Text Carefully**: Pay close attention to the provided section of the text to fully understand the narrative context, including any historical or cultural references.
      
      2. **Check for Irrelevant Information**: Identify whether the text contains non-narrative elements such as publisher details, web content, disclaimers, or any information not directly related to the narrative. If such content is found, return 'no questions extracted.'
      
      3. **Identify Key Narrative and Contextual Elements**: If the text is free from irrelevant information, focus on identifying the key events, actions, characters, settings, and any historical or cultural references. Consider what is happening, who is involved, where and when these events are taking place, and the historical or symbolic significance of these elements.
      
      4. **Formulate Questions**: Create questions that are specific to the identified narrative and contextual elements. Ensure each question is concise, detailed, factual, and directly connected to the content of the narrative, including its historical, cultural, or symbolic context.
      
      5. **Question Variety and Depth**: Aim for a diverse set of questions that cover various aspects of the narrative, including specific locations, character roles, relationships, and cultural or historical context. Avoid redundancy by ensuring each question explores a different element or angle of the narrative.
      
      6. **Avoid Abstract and Meta-Content**: Refrain from generating questions about abstract themes, philosophical ideas, or meta-information such as publication details or background information unrelated to the narrative itself.
    \newline
      
      **Example**:
    \newline
      
      Here is an excerpt from the book:
      
      ---
            \newline
\"The Great Peace towards which people of good will throughout the centuries have inclined their hearts, of which seers and poets for countless generations have expressed their vision, and for which from age to age the sacred scriptures of mankind have constantly held the promise, is now at long last within the reach of the nations. For the first time in history it is possible for everyone to view the entire planet, with all its myriad diversified peoples, in one perspective. World peace is not only possible but inevitable. It is the next stage in the evolution of this planet—in the words of one great thinker, 'the planetization of mankind'. Whether peace is to be reached only after unimaginable horrors precipitated by humanity’s stubborn clinging to old patterns of behaviour, or is to be embraced now by an act of consultative will, is the choice before all who inhabit the earth. At this critical juncture when the intractable problems confronting nations have been fused into one common concern for the whole world, failure to stem the tide of conflict and disorder would be unconscionably irresponsible." 

      ---
      \newline
      
      **Example Questions**:
      \newline
      
      - Where is the Great Peace expected?
      
      - Who has expressed the vision of the Great Peace?
      
      - What does 'planetization of mankind' mean?
      
      - How does the text describe the current world state?
      
      - What critical choice is presented?
      \newline
      
      **Your Turn**:
      \newline
      
      Now, using the provided section of text, check for any irrelevant information. If you find any, return 'no questions extracted.' If not, generate a list of specific, concise questions covering various narrative elements such as characters, actions, settings, historical or cultural references, and symbolic meanings.
      \newline
      ---
      \newline
      
      *Section of the book*
      \newline
      
      [INSERT EXCERPT HERE]" \\

\bottomrule
\end{tabular}
\caption{Prompts for generating speculative questions on NarrativeQA and QuALITY.}
\label{tab: ques_spec_prompts_v1}
\vspace{-5pt}
\end{table*}

\begin{table*}[h!]
\centering
\small
\begin{tabular}{@{}p{0.9\textwidth}}
\toprule
\textbf{Qasper (Question Speculation)} \\
\midrule
\textbf{System: }You are an AI language model that generates insightful and analytical questions about a given passage. Your goal is to create questions that encourage deeper understanding and critical thinking about the content, themes, and details within the passage. The questions should resemble the style of the example questions provided.
\newline

\textbf{User: }

**Instructions:**

1. Carefully read the passage provided, paying special attention to any mention of the experimental design, dataset details, evaluation methods, and results.

2. Generate a list of questions focusing on the following aspects:
   - Experimental setup
   - Dataset characteristics (e.g., size, composition)
   - Evaluation methods and metrics
   - Results and conclusions

3. The questions should be clear, specific, and thought-provoking, encouraging a deep understanding of the methodology and results presented.

4. **Each question must contain only one question.**

5. **Extract as many questions as possible.**
\newline

**Example:**

\_Passage:\_

"Minimally Supervised Learning of Affective Events Using Discourse Relations

Recognizing affective events that trigger positive or negative sentiment has a wide range of natural language processing applications but remains a challenging problem mainly because the polarity of an event is not necessarily predictable from its constituent words. In this paper, we propose to propagate affective polarity using discourse relations. Our method is simple and only requires a very small seed lexicon and a large raw corpus. Our experiments using Japanese data show that our method learns affective events effectively without manually labeled data. It also improves supervised learning results when labeled data are small.

Introduction

Affective events are events that typically affect people in positive or negative ways. For example, getting money and playing sports are usually positive to the experiencers; catching cold and losing one's wallet are negative. Understanding affective events is important to various natural language processing (NLP) applications such as dialogue systems, question-answering systems, and humor recognition. In this paper, we work on recognizing the polarity of an affective event that is represented by a score ranging from $-1$ (negative) to 1 (positive).

Learning affective events is challenging because, as the examples above suggest, the polarity of an event is not necessarily predictable from its constituent words. Combined with the unbounded combinatorial nature of language, the non-compositionality of affective polarity entails the need for large amounts of world knowledge, which can hardly be learned from small annotated data.

In this paper, we propose a simple and effective method for learning affective events that only requires a very small seed lexicon and a large raw corpus. As illustrated in Figure 1, our key idea is that we can exploit discourse relations to efficiently propagate polarity from seed predicates that directly report one's emotions (e.g., “to be glad” is positive). Suppose that events $x_1$ are $x_2$ are in the discourse relation of Cause (i.e., $x_1$ causes $x_2$). If the seed lexicon suggests $x_2$ is positive, $x_1$ is also likely to be positive because it triggers the positive emotion. The fact that $x_2$ is known to be negative indicates the negative polarity of $x_1$. Similarly, if $x_1$ and $x_2$ are in the discourse relation of Concession (i.e., $x_2$ in spite of $x_1$), the reverse of $x_2$'s polarity can be propagated to $x_1$. Even if $x_2$'s polarity is not known in advance, we can exploit the tendency of $x_1$ and $x_2$ to be of the same polarity (for Cause) or of the reverse polarity (for Concession) although the heuristic is not exempt from counterexamples. We transform this idea into objective functions and train neural network models that predict the polarity of a given event.

chatWe trained the models using a Japanese web corpus. Given the minimum amount of supervision, they performed well. In addition, the combination of annotated and unannotated data yielded a gain over a purely supervised baseline when labeled data were small."

\_Example Questions:\_

1. What is the seed lexicon?

2. How are relations used to propagate polarity?

3. How does their model learn using mostly raw data?

4. How big is the Japanese data?

5. How large is the raw corpus used for training?

6. How big is the seed lexicon used for training?

7. What are the results?

8. What are the labels available in the dataset for supervision?

9. How significant are the improvements of supervised learning results trained on smaller labeled data enhanced with the proposed approach compared to the basic approach?

---

**Task:**

Now, read the following passage and generate a list of questions that resemble the style of the example questions.

\_Passage:\_

[INSERT EXCERPT HERE]\\

\bottomrule
\end{tabular}
\caption{Prompt for generating speculative questions on Qasper.}
\label{tab: ques_spec_prompts_v2}
\vspace{-5pt}
\end{table*}

\begin{table*}[h!]
\centering
\small
\begin{tabular}{@{}p{0.9\textwidth}}
\toprule
\textbf{NarrativeQA \& QuALITY (KB Construction)} \\
\midrule
\textbf{System}: You are a helpful assistant.
\newline
\textbf{User}: Please extract all relevant entities and facts from the provided passage that are useful for answering specific questions. Only return entity and facts for information that is explicitly mentioned in the passage. If a question does not have a corresponding fact in the passage, omit that entity and fact entirely. For example, if the question is "Who visits the philosopher at the beginning of the story?" and the passage mentions that a friend visits the philosopher, the response should be (Visitor, A friend visits the philosopher). However, if the passage does not provide specific information on a question and there is no mention of the location, do not include anything in your response for that question. Your returned output should be a series of tuples, like (Visitor, A friend visits the philosopher), (Philosopher's stance on law, Breaking the law is equivalent to betraying a contract with the state).

Passage:
[INSERT EXCERPT HERE]

Questions:
[INSERT SPECULATED QUESTIONS HERE]
 \\

\bottomrule
\end{tabular}
\caption{Prompt for constructing knowledge bases using speculative questions from NarrativeQA or QuALITY.}
\label{tab: kb_prompts_v1}
\end{table*}

\begin{table*}[h!]
\centering
\small
\begin{tabular}{@{}p{0.9\textwidth}}
\toprule
\textbf{Qasper (KB Construction)} \\
\midrule
\textbf{System}: You are a helpful assistant.
\newline
\textbf{User}: Please provide answers to the following questions based on the passage. Whenever possible, prioritize using **direct quotes** from the passage instead of summarizing. Only summarize when a direct quote does not provide a clear answer. Format each answer as a pair of:

(Question, Answer)

If a direct quote is used, place it within quotation marks.

Example format:

(What is the seed lexicon?, A vocabulary of positive and negative predicates that helps determine the polarity score of an event.)

(How big is the Japanese data?, 7,000,000 pairs of events were extracted from the Japanese Web corpus, and 529,850 pairs of events were extracted from the ACP corpus.)

(How does the proposed method compare to previous techniques?, "Compared to existing methods, the proposed approach 'achieves a 15\% increase in classification accuracy while reducing computational complexity by approximately 30\%.' This substantial improvement highlights the efficiency and effectiveness of the new algorithm in large-scale data settings.")

Passage:
[INSERT EXCERPT HERE]

Questions:
[INSERT SPECULATED QUESTIONS HERE]
 \\

\bottomrule
\end{tabular}
\caption{Prompt for constructing knowledge bases using speculative questions from Qasper.}
\label{tab: kb_prompts_v2}
\end{table*}

\begin{table*}[h!]
\centering
\small
\begin{tabular}{@{}p{0.9\textwidth}}
\toprule
\textbf{NarrativeQA (Question Answering - round 1)} \\
\midrule
\textbf{System}: You are a helpful assistant.
\newline
\textbf{User}: 
Please answer the question below using the provided context. Your response must be a phrase that directly answers the question or the phrase 'I don't know'—no further explanation should be added. Do not provide additional context or clarification in your response. Keep the replies concise and short. Do not repeat things. Do not over-explain yourself. Reply in under 10 words.

Example 1:

Context: [(the morning star, The entity known as 'the morning star' is also referred to by another name in astronomy.)]

Question: What is another name for the morning star?

Answer: Venus.

Example 2:

Context: [(The battle of Hastings, The battle of Hastings was fought in the year 1066.)]
Question: When was the battle of Hastings fought?
Answer: 1066.

Example 3:

Context: [(the foundational document, The document foundational to the laws of the United States is the Constitution.)]

Question: What is the foundational document of the United States?

Answer: The Constitution.

Please answer the question below using the provided context. Your response must be either a phrase that directly answers the question or the phrase 'I don't know'—no further explanation should be added. Do not provide additional context or clarification in your response.

Context: [INSERT RETRIEVED DOCUMENTS HERE], 
Question: [INSERT QUESTION HERE]
 \\

\bottomrule
\end{tabular}
\caption{Prompt for answering questions from Qasper.}
\label{tab: qa_prompts_v1}
\end{table*}

\begin{table*}[h!]
\centering
\small
\begin{tabular}{@{}p{0.9\textwidth}}
\toprule
\textbf{NarrativeQA (Question Answering - round 2)} \\
\midrule
\textbf{System}: You are a helpful assistant.
\newline
\textbf{User}: 
For the question-answer pair provided below, shorten the answer by removing any redundant elements that merely repeat information from the question. Only shorten the answer if it includes unnecessary details or redundant phrasing, ensuring that all essential information is retained. Use these provided examples as a guide for the style and level of conciseness expected in the responses.

Examples:

1. **Question:** Who was Socrates visited by at the beginning of the story?

    - **Original Answer:** I don't know. The context provided does not mention anyone visiting Socrates at the beginning of the story.
    
    - **Shortened Answer:** I don't know.
    
2. **Question:** What does Socrates tell Crito not to worry about?

    - **Original Answer:** Socrates tells Crito not to worry about the voices of the crowd regarding Socrates' choices, and not to concern himself with the fairness of the laws.
    
    - **Shortened Answer:** The voices of the crowd.
    
3. **Question:** Who announces the events that are to come to the dismay of the others on stage?

    - **Knowledge Base:** The character who announces the events that are to come; Identity, Phantastes.
    
    - **Shortened Answer:** Phantastes.
    
4. **Question:** Where do the dancers purify themselves?

    - **Original Answer:** In the temple of Apollo.
    
    - **Shortened Answer:** In the temple of Apollo.
    
5. **Question:** Where is Echo's glade?

    - **Original Answer:** Echo's glade is in the forest of Arden.
    
    - **Shortened Answer:** Arden.
    
6. **Question:** What challenge does Phronimus propose to all comers?

    - **Original Answer:** Phronimus proposes a wit duel to all comers.
    
    - **Shortened Answer:** Wit duel.
    
7. **Question:** How long has Michael lived in New York?

    - **Original Answer:** Michael has lived in New York for fifteen years.
    
    - **Shortened Answer:** Fifteen years.
    
8. **Question:** Who wins the sparring match between Johnny and Tom?

    - **Original Answer:** Tom wins the sparring match between Johnny and Tom.
    
    - **Shortened Answer:** Tom.

**Question:** [INSERT QUESTION HERE]

    - **Original Answer:** [INSERT ANSWER FROM ROUND 1]
    
    - **Shortened Answer:** 

Context: [INSERT RETRIEVED DOCUMENTS HERE], 
Question: [INSERT QUESTION HERE]
 \\

\bottomrule
\end{tabular}
\caption{Prompt for answering questions from Qasper.}
\label{tab: qa_prompts_v2}
\end{table*}

\begin{table*}[h!]
\centering
\small
\begin{tabular}{@{}p{0.9\textwidth}}
\toprule
\textbf{Qasper (Question Answering)} \\
\midrule
\textbf{System}: You are a helpful assistant.
\newline
\textbf{User}: 
**Instructions:**

    1. If you find direct evidence from the context, extract the relevant span as your answer. Ensure it is concise and faithful to the text.
    
    2. If the answer requires a rephrasing or cannot be directly extracted, use your own words to provide a clear, concise response.
    
    3. For yes/no questions, simply respond with 'Yes' or 'No' based on the context.
    
    4. If no answer is found within the context, output 'Unanswerable.'

    **Context:**
    [INSERT RETRIEVED DOCUMENTS HERE]

    **Question:**
    [INSERT QUESTION HERE]
 \\

\bottomrule
\end{tabular}
\caption{Prompt for answering questions from Qasper.}
\label{tab: qa_prompts_v3}
\end{table*}

\begin{table*}[h!]
\centering
\small
\begin{tabular}{@{}p{0.9\textwidth}}
\toprule
\textbf{QuALITY (Question Answering)} \\
\midrule
\textbf{System}: You are a helpful assistant.
\newline
\textbf{User}: Please answer the following multiple-choice question based on the context provided.

**Context:**
[INSERT EXCERPT HERE]

**Question:**
[INSERT QUESTION HERE]

**Options:**
1. {options[0]}
2. {options[1]}
3. {options[2]}
4. {options[3]}

Choose the option that seems most appropriate based on the context, even if you're unsure. Respond with only the number of the selected option and do not provide any additional text or explanation.
 \\

\bottomrule
\end{tabular}
\caption{Prompt for answering questions from QuALITY.}
\label{tab: qa_prompts_v4}
\end{table*}

\end{document}